\documentclass[runningheads]{llncs}

\usepackage{graphicx}
\usepackage{tikz}
\usepackage{hyperref}
\usepackage{cite}
\usepackage{multirow}
\usepackage[long]{iaclpaper}

\graphicspath{{imgs/}}

\usetikzlibrary{fit}
\usetikzlibrary{shapes.misc}
\usetikzlibrary{shapes.geometric}

\title{A Structural Causal Model for \\ MR Images of Multiple Sclerosis\Long{\thanks{This research was supported by the NIH (R01- NS082347, PI: Peter Calabresi)}}}
\author{Jacob~C.~Reinhold\inst{1} \and
        Aaron~Carass\inst{2} \and
        Jerry~L.~Prince\inst{1,2}}
\authorrunning{J.~Reinhold et al.}
\institute{
Department~of~Electrical~and~Computer~Engineering, 
Johns~Hopkins~University, Baltimore,~MD,~USA~21218\\
\email{jcreinhold@gmail.com,\{aaron\_carass,prince\}@jhu.com} \and
Department~of~Computer~Science, 
Johns~Hopkins~University, Baltimore,~MD,~USA~21218
}

\begin{document}
\maketitle
\begin{abstract}
Precision medicine involves answering counterfactual questions such as
``Would this patient respond better to treatment A or treatment B?''
These types of questions are causal in nature and require the tools of
causal inference to be answered, e.g., with a structural causal
model~(SCM). In this work, we develop an SCM that models the interaction
between demographic information, disease covariates, and magnetic
resonance~(MR) images of the brain for people with multiple
sclerosis. Inference in the SCM generates counterfactual images
that show what an MR image of the brain would look like if 
demographic or disease covariates are changed. These images can be
used for modeling disease progression or used for image
processing tasks where controlling for confounders is necessary.
\end{abstract}
\section{Introduction}
\Long{Scientific inquiry and precision medicine involve answering causal questions, 
e.g., ``does this medicine treat this disease?'' or ``does this immune response 
cause this symptom?'' Causal questions are asked to determine the effect of 
interventions on variables of interest. The main tool scientists use 
to investigate phenomena, however, is statistical inference which lacks effective 
methods to establish causal relationships outside of randomized control trials~(RCTs).}
\Short{Scientific inquiry and precision medicine involve answering causal
questions, e.g., ``does this medicine treat this disease?''
Causal questions are asked
to determine the effect of interventions on variables of interest. The
main tool used to investigate these phenomena, however, is
statistical inference which lacks effective methods to establish
causal relationships outside of randomized control trials~(RCTs).}

\Long{
The field of causal inference broadens the number of tools scientists have
to establish causal effect. If the scientist has a plausible model
of the relationships between covariates, and the proper measurements 
are made such that confounders are controlled for, then causal 
questions\textemdash like the ones at the beginning of this 
paper\textemdash can be answered with observational data instead of an RCT
\cite{pearl2009causality,peters2017elements,morgan2015counterfactuals}.
A convenient way for scientists to express their prior knowledge about 
relationships between covariates is with a directed acyclic graph (DAG) 
where the edges point from cause to effect. The edge reflects the assumption
that the effect is a function of the cause and potentially some noise 
that accounts for uncertainty. When a DAG has such causal interpretations
it is called a structural causal model (SCM), and it represents a generative 
model of the data on which we can emulate interventions and generate counterfactuals.}
\Short{If a scientist has a plausible model of the relationships between
covariates, and proper measurements have been made (controlling for 
confounds), then causal questions can be answered with observational
data instead of an RCT~\cite{pearl2009causality, peters2017elements}.
A directed acyclic graph~(DAG) can serve as such a model, with edges 
pointing from cause to effect, i.e, the effect is a function of the cause.
Such a DAG is known as a structural causal model~(SCM), representing a generative 
model on which we can emulate interventions and generate counterfactuals.}

\Long{
In medical image analysis, many research questions are naturally 
about counterfactuals \cite{castro2020causality}. For example, image harmonization can be viewed as asking
the counterfactual question: ``What would this image look like if it had been 
acquired with scanner X at site A?'' Super-resolution can be viewed as asking 
the question: ``What would this image look like if it had been acquired 
with 1 cm\textsuperscript{3} resolution?''}
\Short{Medical imaging problems like image harmonization can be viewed as
asking the counterfactual question: ``What would this image
look like if it had been acquired with scanner X?'' \cite{castro2020causality}.
Inference in SCMs, however, is difficult for such high-dimensional
problems. A tractable approach is to amortize inference across datasets and
local variables with a neural network~\cite{gershman2014amortized,
ritchie2016deep}. Pawlowski~et~al.~\cite{pawlowski2020dscm} 
implemented an SCM with amortized inference for healthy MR images of the brain; 
we extended their model to the clinical and radiological phenotype of multiple
sclerosis~(MS)~\cite{reich2018ms}. MS presents as hyper-intense lesions in
$T_2$-weighted MR images, which are readily apparent in white
matter~(WM).
}

\Long{Inference in SCMs, however, is difficult for high-dimensional problems
like medical images. Even approximate methods like variational inference 
\cite{jordan1999introduction,blei2017variational} 
don't scale to high-dimensional problems well. A more tractable approach 
is to amortize the inference across datasets (and local variables) with a 
neural network \cite{gershman2014amortized,ritchie2016deep}. Pawlowski et al. 
\cite{pawlowski2020dscm} used such methods to implement a SCM for 
healthy MR images of the brain, and we extended their model to account
for the clinical and radiological phenotype of multiple sclerosis (MS).}

\Long{MS is an autoimmune disease with a typical onset in early adulthood that 
affects more than two million people worldwide \cite{reich2018ms}. In $T_2$-w 
structural MR images of the brain, MS presents as hyper-intense lesions where 
neurons have demylinated. The most readily apparent lesions are in the white 
matter, although demylinated lesions can be located in any neural tissue.}

\Long{In this paper, we propose an SCM that encodes a causal functional relationship
between demographic and disease covariates with MR images of the brain
for people with MS. Such a model allows us to answer 
counterfactual questions like: ``What would this subject's brain image look 
like if the subject did not have lesions given that they have a 60 mL lesion 
load?'' and ``What would this subject's brain image look like if the subject 
had six years of symptoms given that they have three years of symptoms?''}

\Long{Answers to such counterfactual questions improve our understanding of the 
relationship between the clinical phenotype and the presentation of disease 
in MR images. Studying this relationship is of particular interest in MS because 
of the ``clinico-radiological paradox,'' which has to do with the moderate 
correlation between lesion burden and cognitive performance in people
with MS \cite{mollison2017clinico}.}

\Long{Our contributions are: 1) an extension of the SCM proposed by Pawlowski et al.~\cite{pawlowski2020dscm}
to include covariates related to MS and 2) a novel method to generate realistic 
high-resolution counterfactual images with the variational autoencoders (VAE) 
embedded in the SCM.}

\Short{We propose an SCM that encodes a causal functional relationship with
MR images of the brain for people with MS (PwMS). With this we can answer 
counterfactual questions like: ``What would this brain look
like if it did not have lesions?'' Counterfactual questions improve
our understanding of the relationship between clinical phenotype
and disease presentation in MR. This is of interest in MS because of the 
``clinico-radiological paradox,'' related to the moderate correlation between 
lesion load and cognitive performance. Our contributions are: 1)~an extension of 
the SCM model~\cite{pawlowski2020dscm} to include MS-related covariates and 
2)~a novel method to generate realistic high-resolution~(HR) counterfactual images 
with a variational autoencoder~(VAE) embedded in the SCM.}

\section{Related work}
\Short{
\begin{figure}[t]
    \centering
    \includegraphics[width=\textwidth]{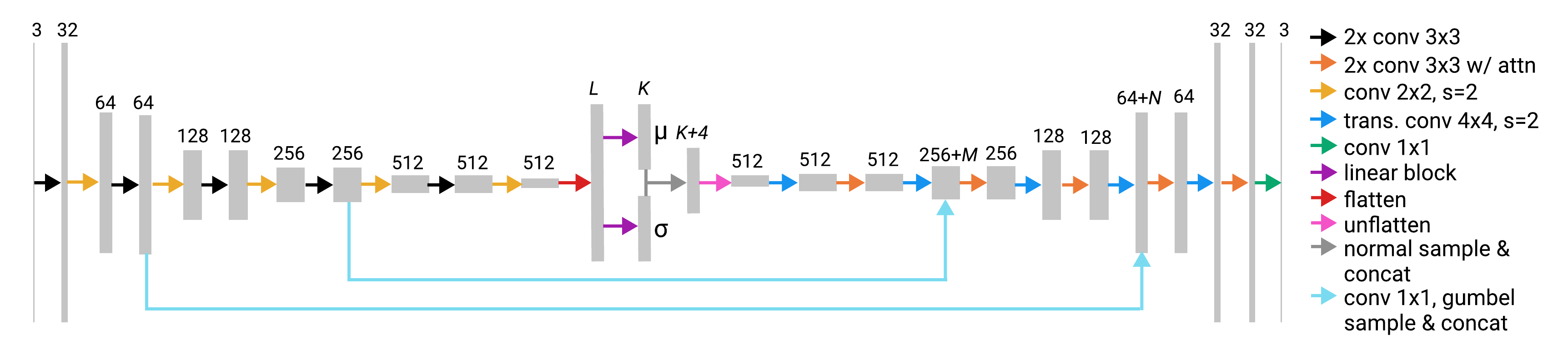}
    \caption{\textbf{VAE with hierarchical latent space}: Deep neural network representing the 
    recognition model~(encoder) and generative model of the observed image~(decoder). The variables 
    $K,L,M,N\in\N$ are experiment-specific parameters.}
    \label{fig:vae}
\end{figure}
}

\Long{The work presented in this paper is most directly related to disease modeling 
and prediction, which is important both to understand disease and to improve 
clinical outcome. Recently, machine learning (ML) has been used to create data-driven 
disease models, e.g., by training a neural network to predict how an MR image of 
the brain of a patient with Alzheimer's will look after a certain amount of time 
\cite{ravi2019degenerative}. The predicted images generated from these methods
may appear convincing, but methods that don't make assumptions about causal 
structure are fundamentally limited. A deep neural network used to predict
disease progression may use spurious correlations in the data to
make predictions that accurately fit the data but do not reflect the true
causal effect \cite{shpitser2008complete}. Traditionally, an RCT is performed
to establish causal effect because observational data contains potential confounders. 
An SCM, by contrast, makes explicit causal assumptions and 
can consequently control for confounders when the appropriate measurements are made, providing a more
sound foundation on which to develop models of disease.}
\Short{Machine learning (ML) has recently been used to create data-driven disease
models, e.g., by training a network to show the MR progression of Alzheimer's~\cite{ravi2019degenerative}.
The images may appear convincing, but methods that don't make assumptions about 
causal structure are fundamentally limited. A deep network trained to predict
disease progression may use spurious correlations that fit the data
but do not reflect the true causal effect~\cite{shpitser2008complete}.
RCTs are traditionally used to establish causal effect because
observational data contains potential confounders. An SCM, by
contrast, makes explicit causal assumptions and can control for
confounders, providing a more robust model of disease.}

\Long{This work also has application to lesion filling for structural MR images of people 
with MS. Lesion filling attempts to make lesions appear 
like healthy tissue\textemdash usually healthy white matter\textemdash for the purpose 
of image processing tasks like whole brain segmentation \cite{zhang2020robust}. 
These lesion-filled images are sometimes referred to as \emph{pseudo-healthy} images \cite{bowles2017brain,xia2020pseudo}.
Lesion filling is useful because image processing methods are usually developed 
on images of people without lesions. The method we propose here can be used for lesion filling
(by intervening to set the lesion volume to 0); however, our method is more general as we can 
generate an image corresponding to any covariate change in the SCM. Our method,
for example, can add lesions to otherwise healthy images (by intervening to set the lesion
volume to some positive value). Our method, as stated previously, also provides a systematic 
way to control for confounders. One approach to lesion filling/pseudo-healthy image synthesis 
is to use a CycleGAN \cite{zhu2017unpaired,xia2020pseudo} where one domain is healthy and the other is MS. Such 
a system, however, can pick up other features of the image that aren't relevant to lesion filling. For 
example, the healthy images in the training set could happen to have relatively large ventricles; 
this could result in the MS-to-healthy mapping making the ventricles larger\textemdash a generally undesired
side effect. This tendency for undesired changes was noted by Cohen et al. \cite{cohen2018distribution}.}
\Short{Our work is also related to WM lesion filling\textemdash where lesions are
made to appear as healthy WM. This is useful because many image processing methods 
are developed on lesion-free images. Our method can be used for lesion filling by 
intervening to set the lesion volume to 0. However, our method is more general as 
we can generate an image for any covariate change in the SCM. It is also more robust 
because it controls for confounders. For example, lesion filling with a CycleGAN, 
where one domain is healthy and the other is MS, may result in undesired side 
effects as noted by Cohen~et~al.~\cite{cohen2018distribution}.}

\Long{
Finally, our work is related to unsupervised learning of disentangled representations \cite{chen2016infogan,higgins2016beta}. 
A disentangled representation is one where at least a subset of the latent variables 
individually control a factor of variation. For example, a disentangled 
representation for images of hand-written digits might include the thickness 
of the digit as one factor of variation, the slant of the digit as another, etc.
Generally, these disentangled factors of variation are implicitly constructed 
to be marginally independent\textemdash an often unrealistic assumption. An example
of factors of variations that are dependent are brain volume and ventricle volume; a 
change in brain volume should be accompanied by an appropriate scaling of ventricle 
volume. The SCM proposed here provides a systematic 
way to intervene on factors of variation\textemdash one that appropriately affects 
downstream dependent factors. The SCM subsumes normal disentangled representations 
because, with the appropriate intervention, you can replicate the interventions 
on factors such that they do not affect other factors. Additionally, disentangled 
representation learning requires an inductive bias \cite{locatello2020sober}, and 
our SCM provides an easily interpretable inductive bias on the factors of variation.
}
\Short{Finally, our work is related to disentangled
representations~\cite{chen2016infogan} which is where a subset of the
latent variables control a factor of variation. These factors are generally 
constructed to be marginally independent which is unrealistic. Our SCM provides a 
systematic way to intervene on factors while accounting for dependency among 
factors. It also efficiently encodes inductive biases necessary to 
disentangle factors~\cite{locatello2020sober}.}

\section{Methods}
\Long{
In this section, we discuss the required background on SCMs, the 
novel VAE component of the SCM that allows our method to scale to 
high-resolution images, and how inference in the SCM is 
implemented within the probabilistic programming language (PPL) Pyro 
\cite{bingham2019pyro}. The code for the model can be found at the
link in the footnote\footnote{\url{https://github.com/jcreinhold/counterfactualms}.}.} 

\subsection{Background on structural causal models}
\Long{An SCM is a tuple~\cite{pearl2009causality}
\begin{equation*} \mathbf{M} = \langle \mathbf{U},\mathbf{V},\mathbf{F},P(\mathbf{u})\rangle, \end{equation*}
where $\mathbf{U}$ are unobserved background variables, $\mathbf{V}$ are 
explicitly modeled variables (called \textit{endogenous} variables), 
$\mathbf{F}$ are the functional relationships between $v_i \in \mathbf{V}$ 
(i.e., $v_i = f_i(\mathrm{pa}_i,u_i)$ where $\mathrm{pa}_i \subset \mathbf{V}$ 
are the parents of $v_i$), and $P(\mathbf{u})$ is the probability distribution 
over $\mathbf{U}$.} 
\Short{An SCM~\cite{pearl2009causality} is a tuple $\bM = \langle
\bU, \bV, \bF, P(\bu) \rangle$, where \bU are unobserved background
variables, \bV are explicitly modeled variables (called
\textit{endogenous} variables), \bF are the functional relationships
between $v_i \in \bV$ (i.e., $v_i = f_i(\mathrm{pa}_i, u_i)$ where
$\mathrm{pa}_i \subset \bV$ are the parents of $v_i$), and $P(\bu)$ is the
probability distribution over \bU.}

\Long{We assume that each element of $\mathbf{U}$ is marginally independent of all others. 
This assumption is referred to as \emph{causal sufficiency} and states that
there is no unobserved confounder \cite{spirtes2010introduction}. We also assume
independence between cause and mechanism; that is, if $\mathrm{pa}_i \to v_i$, then the 
distribution $p(\mathrm{pa}_i)$ and the function $f_i$ are independent \cite{daniusis2012inferring}. 
The upshot is that $f_i$ doesn't change if $p(\mathrm{pa}_i)$ does (e.g., when we do an intervention). 
Finally, we assume $f_j$ is invariant to a change in $f_k$ for $k\ne j$.}
\Short{Each element of \bU is assumed to be independent. This assumption is known as
\emph{causal sufficiency} and also implies \bV includes all
common causes between pairs in \bV~\cite{spirtes2010introduction}. We also assume 
independence between cause and mechanism; that is, if $\mathrm{pa}_i \to v_i$, then the 
distribution of $\mathrm{pa}_i$ and the function $f$ mapping $\mathrm{pa}_i$ to 
$v_i$ are independent of one another \cite{daniusis2012inferring}. This implies
$f_i$ does not change if the distribution of a parent changes. We also assume
$f_j$ is invariant to changes to $f_k$, where $k\ne j$.}

Counterfactual inference is accomplished with three steps:
\Long{\begin{itemize}[leftmargin=5.6em,labelsep=1em,rightmargin=1.5em]
    \item[\textit{Abduction}]Predict all $u_i \in \mathbf{U}$ given
    the functional relationships $f_i \in \mathbf{F}$ and observed $v_i \in \mathbf{V}$.
    \item[\textit{Action}] Modify the graph $\mathbf{M}\to \mathbf{M}_A$ 
    according to the desired intervention (i.e., remove edges going into 
    intervened upon variables), where the intervention is denoted with
    $\textrm{do}(A)$.
    \item[\textit{Prediction}] Use $\mathbf{M}_A$ and the predicted
    $u_i$ to recompute $\tilde{v}_i$ under the functional relationships 
    $f_i$, where $\tilde{v}_i$ are the counterfactuals.
\end{itemize}}
\Short{
\begin{itemize}[leftmargin=5.5em]
    \item[\textit{Abduction}]\quad Predict $u_i, \forall u_i \in \bU$
    given $f_i \in \bF$ and $v_i \in \bV$.
    \item[\textit{Action}]\quad Modify $\bM \to \bM_{A}$ (edge removal) 
    for intervention \DO{A}.
    \item[\textit{Prediction}]\quad Use $\bM_A$ and predicted
    $u_i$'s, to recompute $\tilde{v}_i$ under $f_i$'s.
\end{itemize}
Inference on medical images is challenging for SCMs for numerous
reasons. One issue is $f_i \in \bF$ must be an invertible
function so the $u_i$'s can be computed.
}
\Long{
As discussed previously, high-dimensional data like medical images are challenging 
inference problems for SCMs; generally, $f_i \in \mathbf{F}$ are required to be
invertible functions given $\mathrm{pa}_i$ so that the $u_i\in\mathbf{U}$ can be computed. 
Invertibility, however, comes at a high-computational cost which is prohibitive 
with high-resolution images. As a result, we decompose the $\mathbf{u}_{\mathbf{x}}\in\mathbf{U}$ 
associated with the image $\mathbf{x}$ into a low- and high-level term where the low-level term 
$\boldsymbol{\varepsilon}_{\mathbf{x}}$ is invertible but the high-level term 
$\mathbf{z}_{\mathbf{x}}$ is not. Specifically, the high-level noise term is computed
via explicit, amortized inference with a VAE as discussed in Pawlowski et al. 
\cite{pawlowski2020dscm}. See their paper for details about how the
VAE is used to enable inference in an SCM with images.}
\Short{Invertibility, however, is computationally costly and prohibitive with
HR images. As a result, we decompose $\bu_{\bx} \in \bU$
(associated with image $\bx$) into invertible and non-invertible terms
$\bve_{\bx}$ and $\bz_{\bx}$, respectively. Specifically, $\bz_{\bx}$
is computed explicitly with a VAE, see~\cite{pawlowski2020dscm} for
details.}

\subsection{High-resolution image generation in variational autoencoders}\label{ss:hr}
\Long{
\begin{figure}[t]
    \centering
    \includegraphics[width=\textwidth]{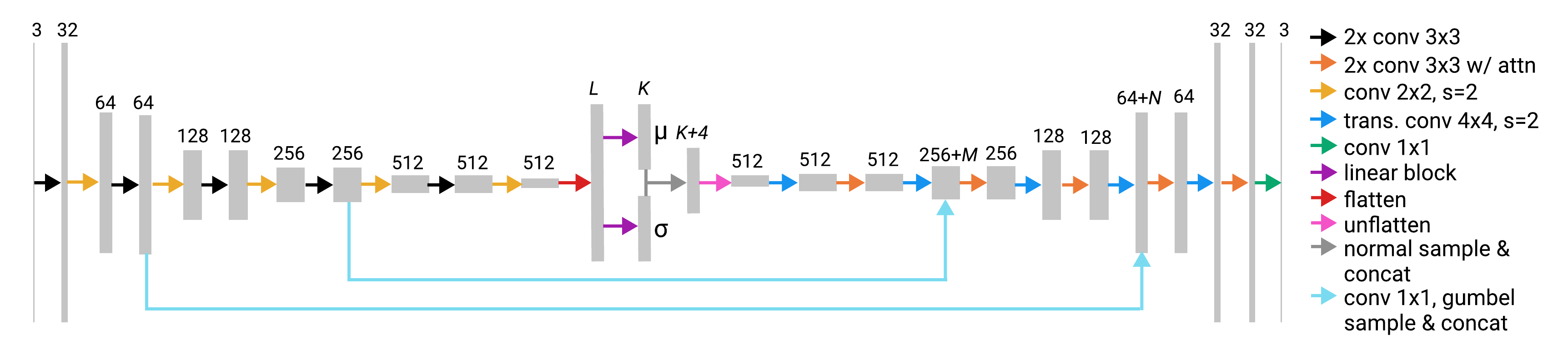}
    \caption{\textbf{VAE with hierarchical latent space}: Deep neural network representing the 
    recognition model~(encoder) and generative model of the observed image~(decoder). The variables 
    $K,L,M,N\in\N$ are experiment-specific parameters.}
    \label{fig:vae}
\end{figure}
}

\Long{As discussed in the previous section, we use a VAE to encode the the high-level 
component of the background variable term associated with the image. 
In Pawlowski et al. \cite{pawlowski2020dscm}, the authors use a simple VAE architecture as a 
proof-of-concept; they used only middle slices of the MR image volumes and 
downsampled them to be low-resolution images ($64\times 64$ pixels). In our work, 
we do two experiments where we use images from a larger range of slices and images 
of size $128\times 128$ and $224 \times 224$ pixels (which are $4\times$ and 
$12.25\times$ larger, respectively).}
\Short{Pawlowski~et~al. \cite{pawlowski2020dscm} use a simple VAE architecture as a proof-of-concept;
they only use the middle slices of an MR image volume downsampled to
$64\times 64$ pixels. In our work, we do two experiments using more
slices and images of size $128\times 128$ and $224 \times 224$ pixels.
HR image generation with VAEs is a topic of increasing
interest~\cite{vahdat2020nvae}. We introduce a novel means of
generating HR images with a VAE by adding a hierarchical binary latent
space~(BLS) $\bz_0$ and $\bz_1$\textemdash in addition to the normal latent space $\bz_2$\textemdash
that preserves structural information. Although inspired by~\cite{dewey2020disentangled}, 
we use a hierarchy of latent spaces and a KL term on the BLS. Our VAE 
architecture is shown in Fig.~\ref{fig:vae} and
a non-counterfactual unconditional sample from
our SCM/VAE is shown in Fig.~\ref{fig:sample}.
}

\Long{High-resolution image generation with VAEs is a topic of increasing interest
\cite{vahdat2020nvae} due to VAEs, in general, being easier to train than GANs. 
In our work, we introduce a novel means of generating high-resolution images
with a VAE by adding a hierarchical binary latent space $\mathbf{z}_0$ and 
$\mathbf{z}_1$\textemdash in addition to the normal latent space $\mathbf{z}_2$\textemdash 
that preserves structural information of the input image without making the VAE 
an identity transformation. This latent space was inspired by the $\beta$ latent space of
Dewey et al. \cite{dewey2020disentangled}. We, however, use a hierarchy of latent
spaces and use a KL term on the binary latent space that isn't present in their work. 
A graphical representation of the VAE architecture is shown in Fig.~\ref{fig:vae}. 
A non-counterfactual, unconditional sample from the SCM with our VAE is shown in Fig.~\ref{fig:sample}.}

\Short{
\begin{figure}[t]
    \centering
    \begin{tabular}{cccccc}
    \includegraphics[width=0.155\textwidth]{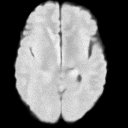}
    \includegraphics[width=0.155\textwidth]{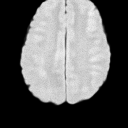}
    \includegraphics[width=0.155\textwidth]{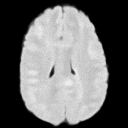}
    \includegraphics[width=0.155\textwidth]{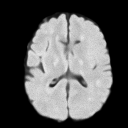}
    \includegraphics[width=0.155\textwidth]{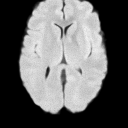}
    \includegraphics[width=0.155\textwidth]{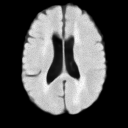}
    \end{tabular}
    \caption{\textbf{Samples from the generative model}: The above six images were
    generated by (unconditionally) ancestrally sampling the generative model trained on 
    $128 \times 128$ images. \Long{The varying location is due to training with random crops.}}
    \label{fig:sample}
\end{figure}

\begin{figure}[b]
    \centering
    \begin{equation}
    \vcenter{\hbox{
    \begin{tikzpicture}[scale=0.70, every node/.style={scale=0.70}]  
    \node[label=$a$,circle,fill,inner sep=2pt,minimum size=1pt] (a) at (0,0) {};
    \node[label=$s$,circle,fill,inner sep=2pt,minimum size=1pt] (s) at (1.5,-.75) {};
    \node[label={[label distance=1pt]360:$e$},circle,fill,inner sep=2pt,minimum size=1pt] (e) at (3,-1.1) {};
    \node[label={[label distance=1pt]270:$\bx$},circle,fill,inner sep=2pt,minimum size=1pt] (x) at (4,-3) {};
    \node[label={[label distance=1pt]360:$n$},circle,fill,inner sep=2pt,minimum size=1pt] (n) at (5,-3) {};
    \node[label={[label distance=1pt]360:$l$},circle,fill,inner sep=2pt,minimum size=1pt] (l) at (4,-2.0) {};
    \node[label=$d$,circle,fill,inner sep=2pt,minimum size=1pt] (d) at (4,0) {};
    \node[label={[label distance=1pt]180:$b$},circle,fill,inner sep=2pt,minimum size=1pt] (b) at (0,-3) {};
    \node[label={[label distance=1pt]180:$v$},circle,fill,inner sep=2pt,minimum size=1pt] (v) at (1.5,-1.75) {};
    \draw[->,thick] (e) -- (l);
    \draw[->,thick] (a) -- (d);
    \draw[->,thick] (d) -- (e);
    \draw[->,thick] (d) -- (v);
    \draw[->,thick] (d) -- (l);
    \draw[->,thick] (l) -- (x);
    \draw[->,thick] (s) -- (d);
    \draw[->,thick] (b) -- (v);
    \draw[->,thick] (a) -- (v);
    \draw[->,thick] (a) -- (b);
    \draw[->,thick] (v) -- (x);
    \draw[->,thick] (b) -- (x);
    \draw[->,thick] (n) -- (x);
    \draw[->,thick] (s) -- (b);
    \draw[->,thick] (s) -- (e);
    \draw[->,thick] (v) -- (l);
    \draw[->,thick] (b) -- (l);
    \end{tikzpicture}
    }}
    \begin{aligned}
      s &= f_s(u_s)       &\qquad    b &= f_b(a, s, u_b) \label{eq:sem}\\
      n &= f_n(u_n)       &          v &= f_v(a, b, u_v) \nonumber \\
      a &= f_a(u_a)       &          l &= f_l(d, e, v, l, u_l) \nonumber \\
      d &= f_d(a, s, u_d) & \bx &= f_{\bx}(b, v, l, n, \bu_{\bx}) \nonumber \\
      e &= f_e(s, d, u_e) \nonumber \\
    \end{aligned}
    \end{equation}
        \caption{\label{fig:pgm}Proposed structural causal model. (Only $\mathbf{V}$ are shown in the
        graph.) $a$ is age, $d$ is the duration of MS symptoms, $l$ is the lesion volume of the subject, 
        $n$ is the slice number, \bx is the image, $b$ is the brain volume, $s$ is biological sex, 
        $e$ is the expanded disability severity score~(EDSS), and $v$ is the ventricle volume. The right-hand
        side shows the functional relationships \bF associated with \bV and \bU of the SCM.}
\end{figure}
}

\Long{The binary latent space consists of a straight-through relaxed Bernoulli 
distribution using the Gumbel reparameterization \cite{jang2017categorical}
on two levels of the feature space. (The locations of which can be seen in Fig.~\ref{fig:vae}
in the light blue lines.) The larger image-dimension binary latent space $\mathbf{z}_0$ is 
$N\times 64 \times 64$ dimensions and the smaller image-dimension binary latent 
space $\mathbf{z}_1$ is $M\times 16 \times 16$ dimension, where $M,N\in\N$ are 
experiment-specific parameters. The prior parameter for the Bernoulli 
distribution for both levels is set to 0.5; the recognition model (encoder) balances minimizing 
the KL divergence of the latent space (i.e., pushing the predicted probability to 0.5) and 
reconstructing the image\textemdash similar to the standard VAE framework with only a normal prior. 
A sample of $\mathbf{z}_0$ is shown in Fig.~\ref{fig:latent}. As can be seen, the binary latent 
space captures a rough outline of the structure of the FLAIR image. It does so without passing 
so much information to the generative model (decoder) that it is unable 
to learn features of the data.}
\Short{The BLS consists of a straight-through relaxed Bernoulli distribution
using the Gumbel reparameterization~\cite{jang2017categorical} on two
levels of the feature space. With $\bz_0$ of size $N \times 64 \times
64$ and $\bz_1$ is $M\times 16 \times 16$, where $M,N\in\N$ are 
experiment-specific parameters. The Bernoulli distribution
has a prior of $0.5$ (for both levels); the encoder balances minimizing
the latent spaces KL divergence and image reconstruction. The BLSs (not shown) are very 
noisy which encourages the network to use all the latent spaces.
}

\Long{As can be seen in Fig.~\ref{fig:latent}, the binary latent spaces are very noisy; this
encourages the network to use the entire hierarchy of latent spaces. An analogy for
why this is the case can be found in Cover and Thomas (Ch. 9) \cite{cover2012elements}.
They discuss communication over multiple parallel noisy channels. The
optimal configuration for maximal information transmission over parallel noisy
channels is to use the channels according to the noise level of each channel 
(e.g., the variance of the noise). We hypothesize that the VAE learns a similar
configuration based on the reliability of information transmission through
the binary latent space.}

\Long{
We used a conditional VAE, conditioned on $\mathbf{c}=(n,b,v,l)^\top$, where $n$ is the slice number, 
$b$ is the brain volume, $v$ is ventricle volume, and $l$ is lesion volume. We concatenate $\bc$ in the normal 
latent space $\mathbf{z}_2$, to make $\mathbf{z}_c=[\mathbf{z}_2,\mathbf{c}]$, which is then
input to the lowest level of the generative model. Additionally, to condition on $\mathbf{c}$ when 
also using binary latent spaces, we used squeeze-and-excite (channel-wise) attention 
\cite{hu2018squeeze} on the convolutional layers indicated with orange arrows in the 
generative model shown in Fig.~\ref{fig:vae}, where the input to the squeeze-and-excite 
layers is $\mathbf{c}$.}

\Short{
We use a conditional VAE, conditioned on $\bc = (n, v ,b, l)^\top$,
where $n$ is slice number, $b$ is brain volume, $v$ is ventricle
volume, and $l$ is lesion volume. We concatenate these values in the
normal latent space $\bz_2$ to make  $\bz_c = [\bz_2, \bc]$, which is
then input to the lowest level of our generative model. To condition
on \bc when also using BLSs, we used squeeze-and-excite (channel-wise)
attention~\cite{hu2018squeeze} on the convolutional layers of the 
generative model, where the input to the squeeze-and-excite layers is \bc.
}
\Long{
\begin{figure}[t]
    \centering
    \begin{tabular}{cccccc}
    \includegraphics[width=0.155\textwidth]{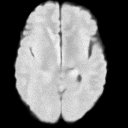}
    \includegraphics[width=0.155\textwidth]{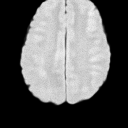}
    \includegraphics[width=0.155\textwidth]{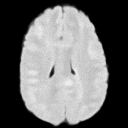}
    \includegraphics[width=0.155\textwidth]{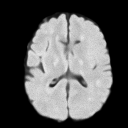}
    \includegraphics[width=0.155\textwidth]{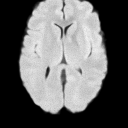}
    \includegraphics[width=0.155\textwidth]{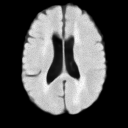}
    \end{tabular}
    \caption{\textbf{Samples from the generative model}: The above six images were
    generated by (unconditionally) ancestrally sampling the generative model trained on 
    $128 \times 128$ images. \Long{The varying location is due to training with random crops.}}
    \label{fig:sample}
\end{figure}
}

\Long{
\begin{figure}[b]
    \centering
    \includegraphics[width=\textwidth]{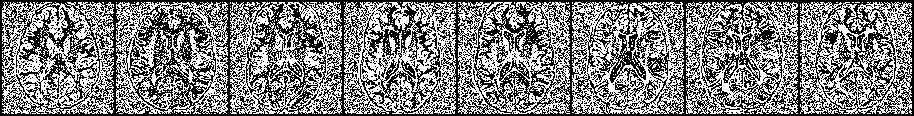}
    \caption{\textbf{Binary latent space}: Visualization of one channel of the binary latent 
    space early in training (later in training it is harder to discern structure).}
    \label{fig:latent}
\end{figure}
}

The generative model outputs image-sized location and scale parameters 
of a Laplace distribution.
For simplicity, we assume pixels are independent so the location and scale images are the same size as the image.
The Laplace distribution was used instead of a Gaussian because the $L_1$ loss 
has been noted to be preferable over MSE for image-related regression tasks~\cite{kendall2017uncertainties}\Short{.}\Long{ and we
noticed a qualitative improvement in the predicted counterfactual images.}

\Long{Training this VAE involves several tricks to stabilize training. We trained the
model for 2000 epochs. We used a one-cycle learning rate schedule \cite{smith2017cyclical} with
the Adam optimizer \cite{kingma2015adam} starting at $2\times 10^{-5}$, increasing to 
$5\times10^{-4}$ for the first 10\% of training epochs, and decaying to $5\times 10^{-8}$. 
We also used a schedule for a scaling parameter $\lambda_i$ on
the KL divergence terms as discussed in S{\o}nderby et al. \cite{sonderby2016ladder};
specifically, we started the KL divergence scaling parameter for the normal distribution $\lambda_2$
at 0 and linearly increased it to a maximum of 1 in 600 epochs. For the binary latent space
KL terms $\lambda_1$ and $\lambda_0$, we started the scaling parameter at a small value and 
linearly increased them to different levels for the smaller and larger latent spaces; these 
settings are experiment-specific. We used weight normalization \cite{salimans2016weight} on 
all convolutional layers. We also clipped the norm of the gradients at a norm of 100. Finally, 
we used one inverse autoregressive affine normalizing flow on the posterior, as discussed by 
Kingma et al. \cite{kingma2016improved}.}
\Short{
We trained the VAE for 2000 epochs using a one-cycle learning rate
schedule~\cite{smith2017cyclical} with the Adam
optimizer~\cite{kingma2015adam} starting at $2\times 10^{-5}$,
increasing to $5\times10^{-4}$ for the first 10\% of training epochs,
and decaying to $5\times 10^{-8}$. We also used a schedule for scaling
the KL terms~\cite{sonderby2016ladder}; specifically, we started the 
KL scaling parameter for the normal distribution, $\lambda_2$, at $0$, and 
linearly increased it $1$ over 600 epochs. The KL terms for the BLSs 
($\lambda_0$ and $\lambda_1$) started at $1$ and linearly increased
them to different levels which were experiment-specific. We
used weight normalization~\cite{salimans2016weight} on all
convolutional layers, and we clipped the norm of the gradients at
100. Finally, we used one inverse autoregressive affine normalizing flow 
on the posterior~\cite{kingma2016improved}.}

\Long{
\subsection{Learning and inference in the SCM and VAE with a PPL}
We used the Pyro probabilistic programming language (PPL) to implement the SCM.
A PPL is a programming language with terms corresponding to sampling and 
conditioning. Pyro is a higher-order\footnote{\emph{Higher-order}
refers to the ability of the programming language to allow general recursion and 
allow functions to be both used as arguments and return values.} PPL built upon 
PyTorch \cite{paszke2019pytorch}. See van de Meent et al. \cite{van2018introduction}
for an overview of PPLs.

The functions $f_i \in mathbf{F}$, which take causal parents $\mathrm{pa}_i$ as input and
outputs some variable $v_i \in \mathbf{V}$, are represented by normalizing flows (NFs)
and neural networks. All $f_i$ in our SCM are linear rational spline NFs 
\cite{dolatabadi2020invertible}, except the functions for the image $f_\mathbf{x}$, biological
sex $f_s$, and slice number $f_n$. The image function $f_\mathbf{x}$ is the generative model part 
of the VAE described in the previous section. Biological sex and slice number are directly sampled from
their base distributions (Bernoulli and uniform, respectively). The base distributions for the 
NFs are normal distributions with the log of the empirical mean and variance of the training 
data as the location and scale parameters, respectively. The base distribution for biological 
sex is a Bernoulli distribution with parameter equal to the mean of data (0 for male, 1 for 
female). The base distribution for slice number is uniform from the minimum to the maximum slice number
in the training set. During training, the hypernetworks associated with the NFs and VAE are jointly 
updated with backpropagation using the (negative) evidence lower bound as a loss function.  
During inference, the learned $\mathbf{F}$ are fixed. A counterfactual image is 
generated using the single-world intervention graph formalism \cite{richardson2013single}.
}
\Short{
\subsection{Learning and inference in the SCM and VAE}
The functions $f_i \in \bF$, which take causal parents $\mathrm{pa}_i$
as input and outputs $v_i \in \bV$, are represented by normalizing
flows~(NFs) and neural networks. All $f_i$ in our SCM are linear
rational spline NFs~\cite{dolatabadi2020invertible}, except the
functions for the image $f_{\bx}$, biological sex $f_s$, and slice
number $f_n$. $f_{\bx}$ is the generative model part of the VAE (see
Sec.~\ref{ss:hr}). The $s$ is sampled from a Bernoulli distribution~($0$ for
male, $1$ for female). The $n$ is sampled from a uniform distribution~(the
minimum to the maximum of slice number in training). The base
distributions for the NFs are normal distributions with the log of the
empirical mean and variance of the training data as the location and
scale parameters, respectively. During training, the networks
associated with the NFs and the VAE are jointly updated with
backpropagation using the negative evidence lower bound as a loss
function. At inference, the learned $\bF$ are fixed. A counterfactual
image is generated using the single-world intervention graph
formalism~\cite{richardson2013single}. The code and a full listing of 
hyperparameters can be found at the link in the 
footnote\footnote{\url{https://github.com/jcreinhold/counterfactualms}}.
}

\section{Experiments}
\subsection{Data}
\Long{
We used a private dataset containing 77 subjects of both healthy control (HC) 
subjects and subjects with MS. We randomly split the data into training, validation, 
and testing. The training set consisted of 68 subjects where 38 were HC
and 30 had MS (29 with had relapsing-remitting MS and 1 had secondary-progressive 
MS). The total number of unique scans in the training set was 171; 124 scans of MS
and 47 HC. The validation set had 7 subjects with 4 HC and 3 MS; 16 total 
unique scans consisting of 5 HC and 11 MS. The test set had 2 subjects with 1 HC and 1 MS;
4 total unique scans consisting of 2 HC and 2 MS. 
}

\Short{We used a private dataset containing 77 subjects of both healthy
controls~(HC) and PwMS. We randomly split the data into training,
validation, and testing. Training consisted of 68~subjects (38~HC,
30~PwMS) which included 171 unique scans~(47~HC, 124~PwMS). The
validation set had 7 subjects~(4~HC, 3~MS) which included 16 unique
scans~(5~HC, 11~MS). The test set had 2 subjects~(1~HC, 1~MS) with 4
unique scans~(2~HC, 2~MS).}
\Long{
Each subject had both a $T_1$-w and a FLAIR image. The $T_1$-w and FLAIR images were
bias field-corrected with N4 \cite{tustison2010n4itk}. The $T_1$-w image was rigidly 
registered to the MNI ICBM152 brain atlas with ANTs \cite{avants2009advanced}. The FLAIR
image was super-resolved with SMORE \cite{zhao2020smore} and then registered
to the $T_1$-w image. The white matter mean was then normalized to 1.0, where the
white matter was found using fuzzy c-means on the $T_1$-w image \cite{reinhold2019evaluating}. 
The brain volume was measured by scaling the number of voxels inside the brain mask, 
found with ROBEX \cite{iglesias2011robust}, according to the digital resolution (0.8 mm\textsuperscript{3}). 
The ventricle volume was similarly computed by segmenting the ventricles with the 
convolutional neural network of Shao et al. \cite{shao2019brain}. The lesion 
volume was measured by training the Tiramisu network described by Zhang et al. 
\cite{zhang2019multiple} on a private dataset. The ventricle segmentation 
occasionally segmented lesions as well as ventricles for subjects with MS; 
the lesion mask was used to remove voxels associated with lesions from the 
ventricle mask. Axial slices of each image were converted to PNGs by 
extracting three adjacent slices, clipping the intensity at the 
99.5\textsuperscript{th} percentile, scaling the image to range between 
$[0, 255]$, and converting the type of the slice array to an 8-bit unsigned 
integer. In all experiments, uniform random noise is added to discrete random 
variables per Theis et al. \cite{theis2016note}.
}
\Short{Each scan had both an MPRAGE and FLAIR image which were bias
field-corrected with N4~\cite{tustison2010n4itk} and the MPRAGE was
registered to the MNI ICBM152 brain atlas with
ANTs~\cite{avants2009advanced}. The FLAIR image was
super-resolved~\cite{zhao2020smore} and then registered to its
corresponding MPRAGE in the MNI space. The WM mean was
then normalized to $1$ using the fuzzy c-means estimated
WM mean~\cite{reinhold2019evaluating}. Brain volume was measured based on
ROBEX~\cite{iglesias2011robust}. Lesion volume was measured based on
Zhang~et~al.~\cite{zhang2019multiple}, trained on a private dataset.
Ventricle volume was computed using Shao~et~al.~\cite{shao2019brain} on
the MPRAGE. The ventricle segmentation occasionally segmented lesions as
ventricles for PwMS; the lesion mask was used to remove the incorrect
voxels and the ventricle volume was updated accordingly. Axial slices of
each image were converted to PNGs by extracting three adjacent
slices, clipping the intensity at the 99.5\textsuperscript{th}
percentile, then rescaling and quantizing to $[0, 255]$. In all experiments, 
uniform random noise is added to discrete random variables~\cite{theis2016note}.
}

\subsection{SCM for MS}

\Long{
\begin{figure}[t]
    \centering
    \begin{equation}
    \vcenter{\hbox{
    \begin{tikzpicture}[scale=0.80, every node/.style={scale=0.80}]  
    \node[label=$a$,circle,fill,inner sep=2pt,minimum size=1pt] (a) at (0,0) {};
    \node[label=$s$,circle,fill,inner sep=2pt,minimum size=1pt] (s) at (1.5,-.75) {};
    \node[label={[label distance=1pt]360:$e$},circle,fill,inner sep=2pt,minimum size=1pt] (e) at (3,-1.1) {};
    \node[label={[label distance=1pt]270:$\bx$},circle,fill,inner sep=2pt,minimum size=1pt] (x) at (4,-3) {};
    \node[label={[label distance=1pt]360:$n$},circle,fill,inner sep=2pt,minimum size=1pt] (n) at (5,-3) {};
    \node[label={[label distance=1pt]360:$l$},circle,fill,inner sep=2pt,minimum size=1pt] (l) at (4,-2.0) {};
    \node[label=$d$,circle,fill,inner sep=2pt,minimum size=1pt] (d) at (4,0) {};
    \node[label={[label distance=1pt]180:$b$},circle,fill,inner sep=2pt,minimum size=1pt] (b) at (0,-3) {};
    \node[label={[label distance=1pt]180:$v$},circle,fill,inner sep=2pt,minimum size=1pt] (v) at (1.5,-1.75) {};
    \draw[->,thick] (e) -- (l);
    \draw[->,thick] (a) -- (d);
    \draw[->,thick] (d) -- (e);
    \draw[->,thick] (d) -- (v);
    \draw[->,thick] (d) -- (l);
    \draw[->,thick] (l) -- (x);
    \draw[->,thick] (s) -- (d);
    \draw[->,thick] (b) -- (v);
    \draw[->,thick] (a) -- (v);
    \draw[->,thick] (a) -- (b);
    \draw[->,thick] (v) -- (x);
    \draw[->,thick] (b) -- (x);
    \draw[->,thick] (n) -- (x);
    \draw[->,thick] (s) -- (b);
    \draw[->,thick] (s) -- (e);
    \draw[->,thick] (v) -- (l);
    \draw[->,thick] (b) -- (l);
    \end{tikzpicture}
    }}
    \begin{aligned}
      s &= f_s(u_s)       &\qquad    b &= f_b(a, s, u_b) \label{eq:sem}\\
      n &= f_n(u_n)       &          v &= f_v(a, b, u_v) \nonumber \\
      a &= f_a(u_a)       &          l &= f_l(d, e, v, l, u_l) \nonumber \\
      d &= f_d(a, s, u_d) & \bx &= f_{\bx}(b, v, l, n, \bu_{\bx}) \nonumber \\
      e &= f_e(s, d, u_e) \nonumber \\
    \end{aligned}
    \end{equation}
    \caption{\label{fig:pgm}\textbf{Proposed structural causal model}: (Only $\mathbf{V}$ are shown in the
        graph.) $a$ is age, $d$ is the duration of MS symptoms, $l$ is the lesion volume of the subject, 
        $n$ is the slice number, \bx is the image, $b$ is the brain volume, $s$ is biological sex, 
        $e$ is the expanded disability severity score~(EDSS), and $v$ is the ventricle volume. The right-hand
        side shows the functional relationships \bF associated with \bV and \bU of the SCM.}
\end{figure}
}

The SCM we use for all experiments is shown as a DAG in Fig.~\ref{fig:pgm}. Note that we only show the endogenous 
variables $\mathbf{V}$ in Fig.~\ref{fig:pgm}. The SCM can alternatively be represented as a system of 
equations relating $\mathbf{V}$ and $\mathbf{U}$ via the equations $\mathbf{F}$ as shown on the right-hand side
of Fig.~\ref{fig:pgm}. The $u_i$ are the background variables associated with an endogenous variable $v_i$. 
Note that $\mathbf{u}_\mathbf{x}=(\mathbf{z}_\mathbf{x},\boldsymbol{\varepsilon}_\mathbf{x})$ where 
$\mathbf{z}_\mathbf{x}=(\mathbf{z}_0,\mathbf{z}_1,\mathbf{z}_2)$ are the non-invertible latent 
space terms estimated with the recognition model of the VAE and $\boldsymbol{\varepsilon}_\mathbf{x}$ 
is the invertible term. The edges in this SCM were determined by starting from the SCM described in 
Pawlowski et al. \cite{pawlowski2020dscm}, densely adding reasonable edges to the new variables 
($d$, $e$, and $l$), and pruning the edges over many experiments based on the quality of the counterfactual images.

\subsection{Small images, large range}

\begin{figure}[t]
    \centering
    \Short{\includegraphics[width=0.95\textwidth]{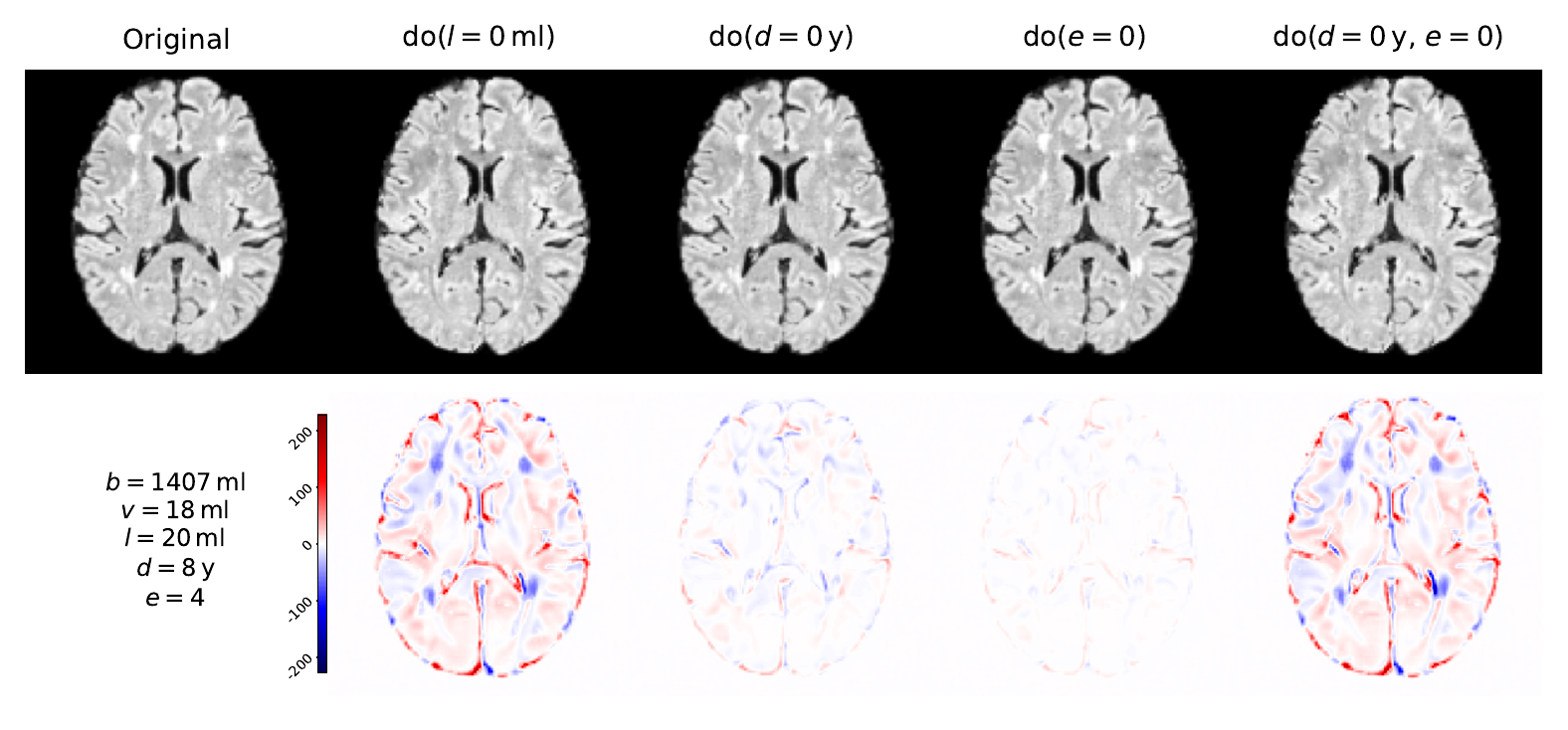}}
    \Long{\includegraphics[width=\textwidth]{small_intervention_1402_lesion_dur_edss_both.pdf}}
    \caption{\textbf{Example counterfactual images}: The first image is the original FLAIR 
    image of a \Short{PwMS}\Long{person with MS}, the remaining images show \Long{an intervention on the
    patient info and the resulting }counterfactual images. From left-to-right the 
    interventions set the 1) lesion volume to 0 mL, 2) duration of symptoms to 0 years, 
    3) EDSS to 0, and 4) duration and EDSS to 0.}
    \label{fig:lesion_interventions}
\end{figure}

\Long{
\begin{figure}[b]
    \centering
    \includegraphics[width=0.95\textwidth]{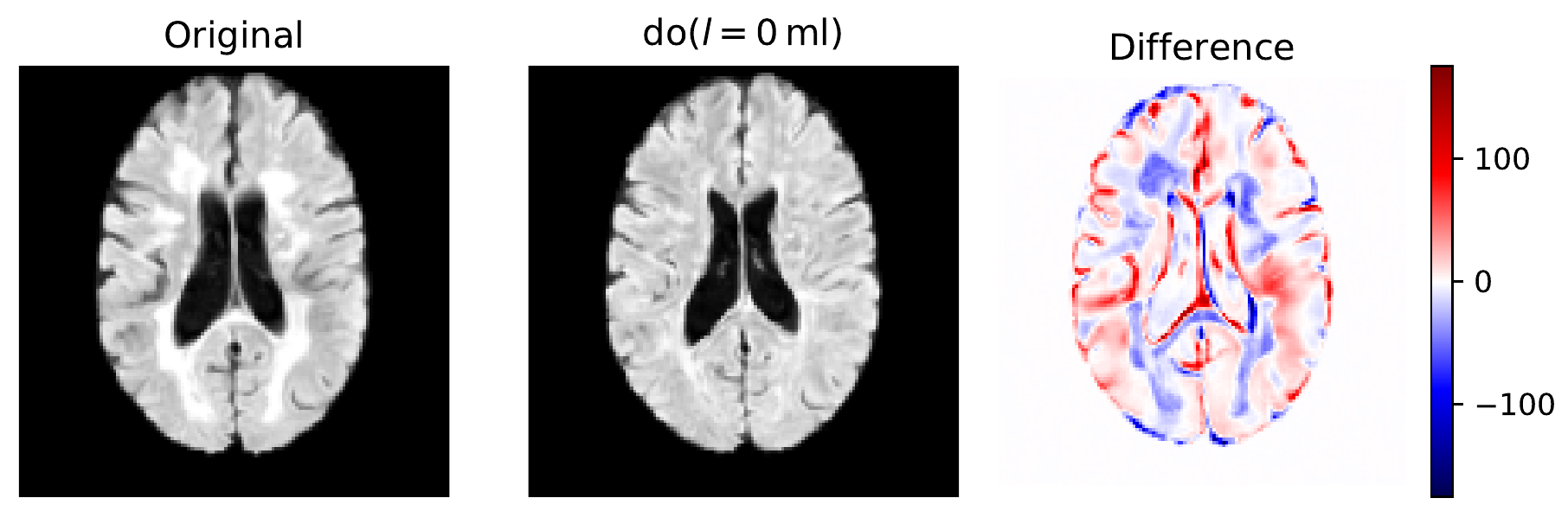}
    \caption{\textbf{Example intervention on lesion volume}: Another example intervention 
    setting lesion volume to 0 mL on subject with a large lesion load (65 mL).}
    \label{fig:extensive_lesion_intervention}
\end{figure}
}

\Long{
In this experiment, we randomly cropped the PNG images to $224\times 224$ pixels and 
downsampled the cropped patches to $128\times 128$ with linear interpolation. We used
the middle 60 axial slices of the image volume as input for training, validation, 
and testing. The KL schedule for the binary latent space $\mathbf{z}_0$ and $\mathbf{z}_1$
both started at $\lambda_0=\lambda_1=1$ and ended at $\lambda_0=4.4$ and $\lambda_1=1.1$. 
We used a batch size of 342 for training. The latent space size parameters are 
$K=100$, $L=8192$, $M=4$, and $N=1$. A more detailed listing of all hyperparameters for 
this experiment can be found at the link in the footnote\footnote{\scriptsize{\url{https://github.com/jcreinhold/counterfactualms/blob/2938c989660d1a6d5aa16cf0c3ae1811bc0fe40b/assets/small_images_large_range.yaml}}}.
}

\Short{
In this experiment, we randomly cropped the PNG images to $224\times 224$ pixels and 
downsampled to $128\times 128$ with linear interpolation. We used 
the middle 60 axial slices of the image volume. The KL schedule for $\mathbf{z}_0$ 
and $\mathbf{z}_1$ started at $\lambda_0=\lambda_1=1$ and ended at $\lambda_0=4.4$ 
and $\lambda_1=1.1$. We used batches of 342 for training. Latent space size parameters are 
$K=100$, $L=8192$, $M=4$, and $N=1$.
}

\Long{
We show an example set of counterfactual images in Fig.~\ref{fig:lesion_interventions}.
Another example counterfactual image is shown in Fig.~\ref{fig:extensive_lesion_intervention}
for a subject with a large lesion load. These counterfactuals are generated from
subjects in the training set; the qualitative quality of counterfactuals on the
validation and test set were worse. Recall, however, that these counterfactuals
are not provided as examples for supervised learning of counterfactual prediction;
predicting counterfactuals is an unsupervised problem. See the discussion section 
for additional details.
}

\begin{figure}[t]
    \centering
    \begin{tabular}{cc}
    \Short{\includegraphics[width=0.45\textwidth]{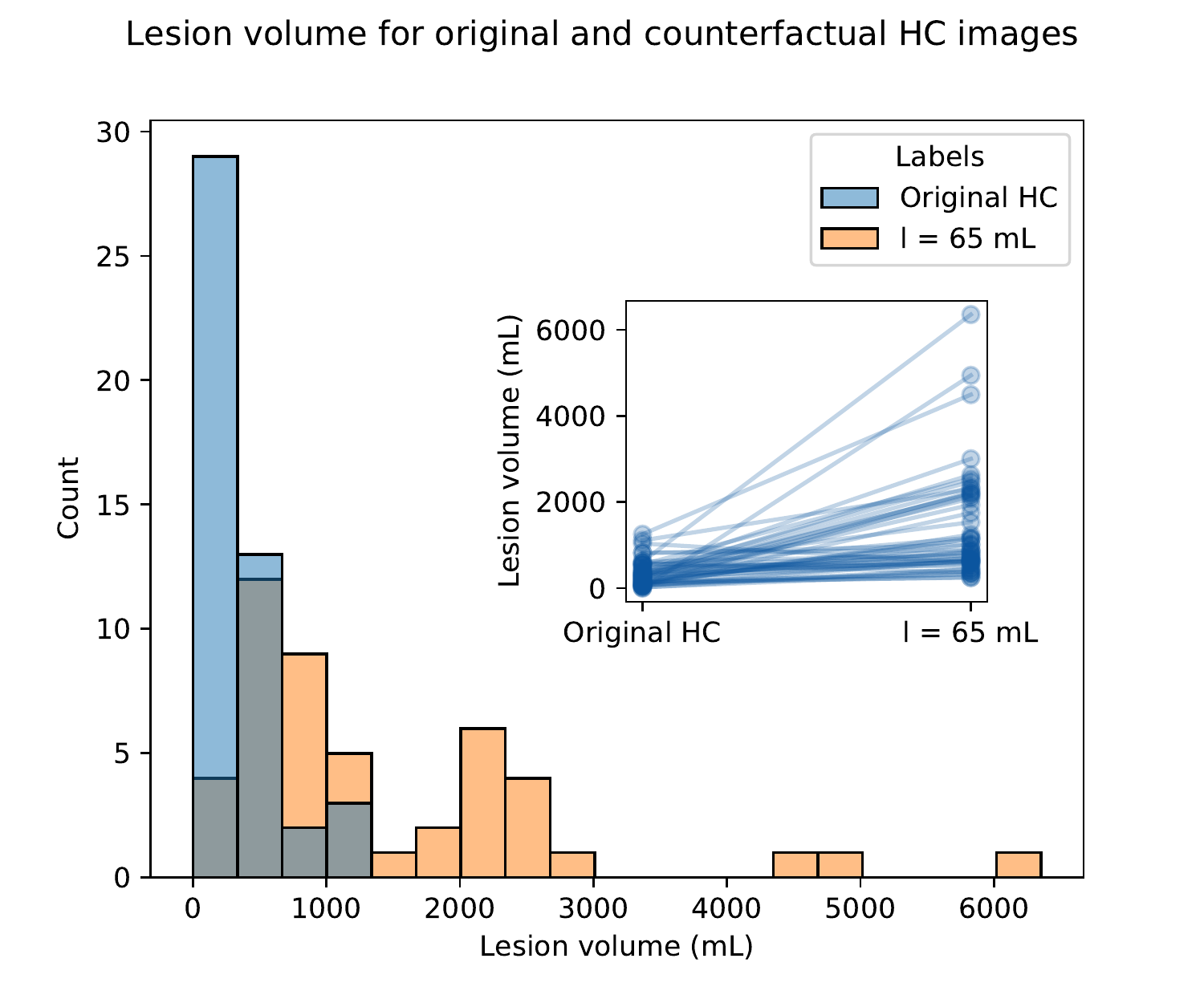} &}
    \Long{\includegraphics[width=0.47\textwidth]{hc_seg_inset.pdf} &}
    \Short{\includegraphics[width=0.45\textwidth]{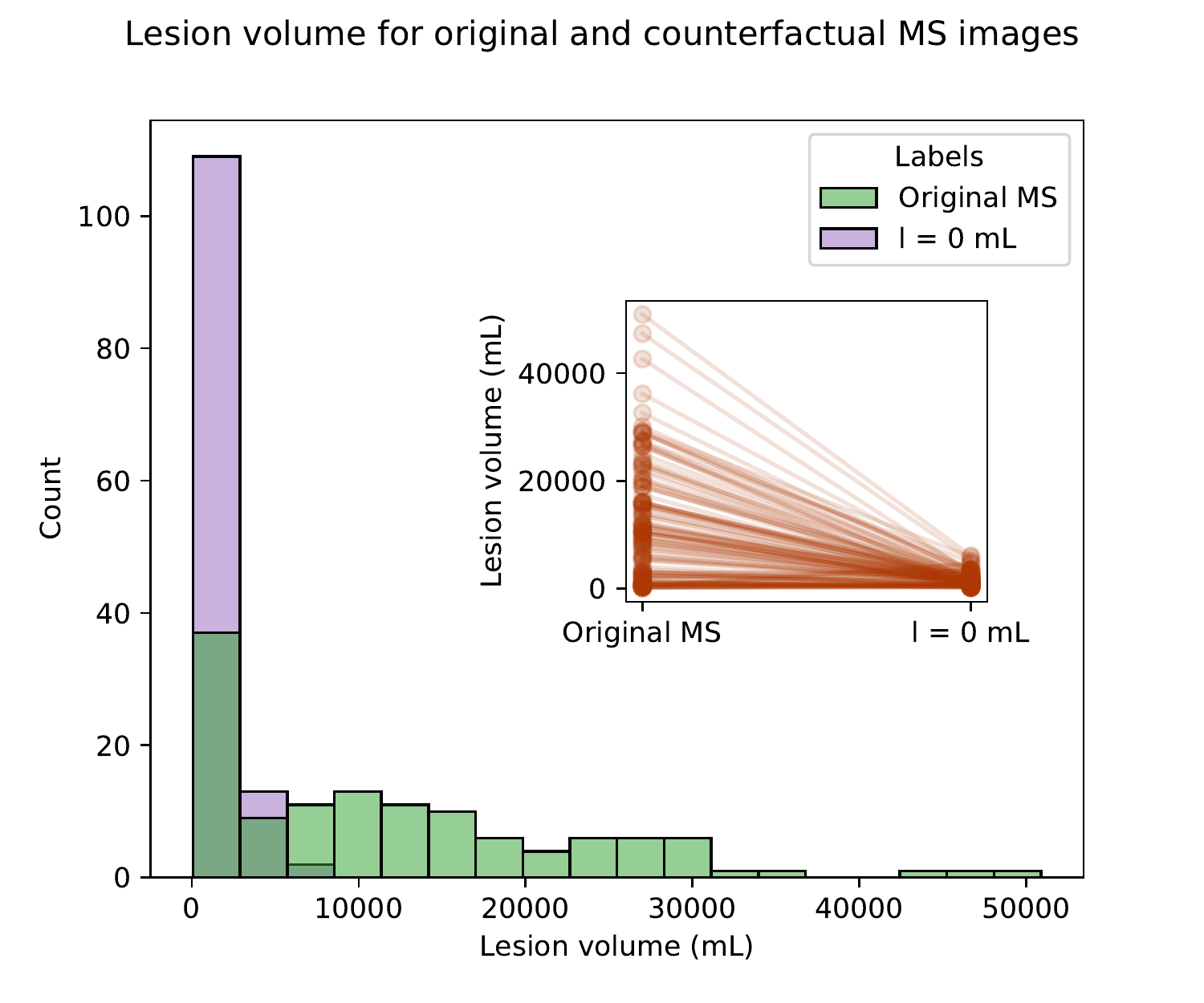}}
    \Long{\includegraphics[width=0.47\textwidth]{ms_seg_inset.pdf}}
    \end{tabular} 
    \caption{\textbf{Lesion segmentation on original and counterfactual images}: The left
    figure shows the histogram of segmented lesion volumes for original and counterfactual HC images,
    where lesion volume was set to 65 mL. The right figure shows a similar plot for original and
    counterfactual MS images, where lesion volume was set to 0 mL. The inset plots show the same
    data with the lesion volume for original on the left and counterfactual on the right.}
    \label{fig:segmentation}
\end{figure}

\Short{
We show an example set of counterfactual images in Fig.~\ref{fig:lesion_interventions}.
These counterfactuals are generated from a subject in the training set; the quality 
of counterfactuals on the validation and test set were worse. See the discussion section 
for details.
}

\Long{
To quantify the effect of our counterfactuals on lesion volume, we 
used a lesion segmentation method on the original and counterfactual images. 
In the first experiment, we segmented HC images before 
and after an intervention setting lesion volume to 65 mL. The results for
this experiment are summarized in the left-hand plot of Fig. \ref{fig:segmentation}.
In the second experiment, we segmented MS images before and after an intervention setting 
lesion volume to 0 mL. The results are in the right-hand plot of the Fig. \ref{fig:segmentation}.
Setting the lesion volume to 0 mL for people with MS consistently moves the lesion volume
to near zero. But for HCs, setting lesion volume to 65 mL does not result in a consistent move to 65 mL.
This is likely due to processing slices individually instead of as a volume. From the perspective
of the SCM, only adding a few lesions in one slice is reasonable because the intervention 
lesion volume can be explained away by the presence of many lesions in other slices. This 
is a limitation of using two-dimensional images to generate counterfactuals with statistics computed
from three-dimensional volumes.
}

\Short{
To quantify the effect of our counterfactuals on lesion volume, we
used a lesion segmentation method on the original and counterfactual images. 
The results are summarized in Fig.~\ref{fig:segmentation}.
Setting the lesion volume to 0 mL for PwMS consistently moves the lesion volume
to near zero; however, for HCs, setting the lesion volume to 65 mL does not result 
in a consistent move to 65 mL. This is likely due to processing slices individually 
instead of as a volume; using 2D images to generate counterfactuals with statistics 
computed from 3D volumes is sub-optimal.
}

\subsection{Large images, small range}
\begin{figure}[t]
    \centering
    \begin{tabular}{cc}
    \begin{tabular}{c}
    \Short{\includegraphics[width=0.65\textwidth]{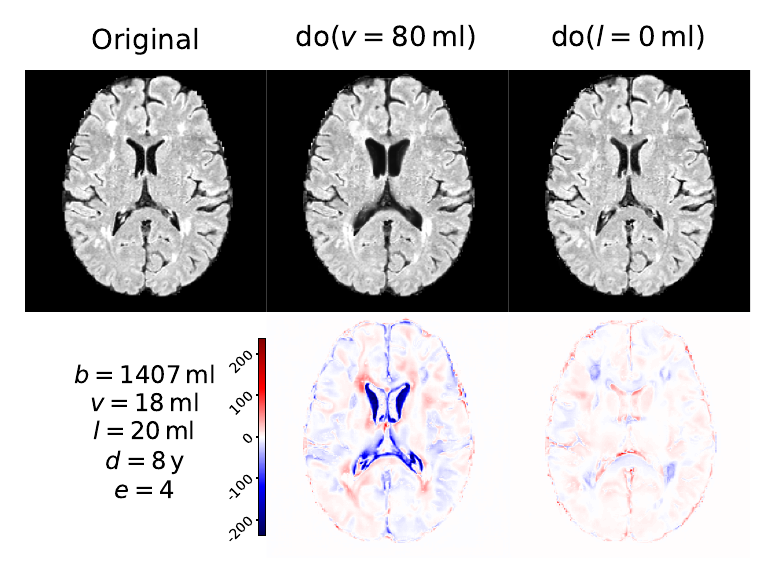}}
    \Long{\includegraphics[width=0.72\textwidth]{large_intervention_1402_lesion_vent.pdf}}
    \end{tabular} &
    \begin{tabular}{c}
    \textsf{Samples} \\
    \Short{\includegraphics[width=0.20\textwidth]{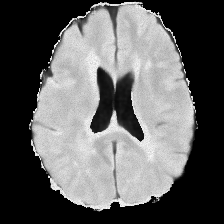} \\}
    \Long{\includegraphics[width=0.22\textwidth]{large_sample_3.png} \\}
    \Short{\includegraphics[width=0.20\textwidth]{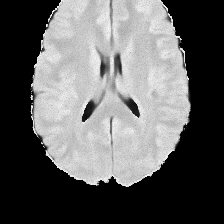}}
    \Long{\includegraphics[width=0.22\textwidth]{large_sample_1.png}}
    \end{tabular}
    \end{tabular}
    \caption{\textbf{Large image counterfactuals and samples}: The images on the left show
    an intervention setting 1) the ventricle volume to 80 mL, 2) the lesion volume to 0 mL. 
    The two images on the right are samples from the generative model, similar to Fig. \ref{fig:sample}.}
    \label{fig:large_sample}
\end{figure}

In this experiment, we cropped the images to $224\times 224$ and did not downsample 
the cropped patches. We used the middle 10 axial slices of the image volume as 
input for training, validation, and testing. The KL warm-up is similar to
the previous experiment except for the KL schedule for $\mathbf{z}_1$ started at $\lambda_1 = 0.5$.
We used a batch size of 128 for training.
The latent space size parameters are $K=120$, $L=25088$, $M=8$, and $N=2$. An
example intervention and samples from the generative model are shown in Fig. \ref{fig:large_sample}.
\Long{A detailed listing of all hyperparameters for this experiment can be found at the link in the footnote\footnote{\scriptsize{\url{https://github.com/jcreinhold/counterfactualms/blob/2938c989660d1a6d5aa16cf0c3ae1811bc0fe40b/assets/large_images_small_range.yaml}}}
}

\section{Discussion}

In this work, we proposed an SCM to generate
counterfactual images with application to MS. We showed that we can produce believable 
counterfactual images given a limited data set of healthy and MS subjects.

\Long{
This work has many limitations. A noted problem with counterfactuals is 
that they are usually unverifiable. Counterfactuals imagine what some variable would look like in a parallel universe 
where all but the intervened on variables and their descendants were the same.
Since we cannot validate our counterfactual images in a meaningful way, 
they should not be directly used for diagnosis or prognosis. The tools of causal 
inference, however, give researchers better ways to control for known confounders. 
As a result, the SCM framework can be potentially improve the performance of and trust in a 
machine learning system, e.g., in a harmonization or synthesis method. 
}
\Short{
This work has many limitations. A noted problem with counterfactuals is 
that they are usually unverifiable. Counterfactuals imagine what some variable 
would look like in a parallel universe where all but the intervened on variables 
and their descendants were the same. Since counterfactual images cannot be validated,
they should not be used for diagnosis or prognosis. The tools of counterfactuals, however, 
give researchers better ways to control for known confounders\textemdash potentially 
improving the performance of and trust in an ML system.
}

Another limitation is that our model created poor counterfactual images for images 
outside the training set. However, causal inference is usually about retrospective 
analysis; generalization to new samples is not a requirement of a successful causal model.
A statistical learning view of ML is that the aim of a predictor 
is to minimize the loss function under the true data distribution (i.e., the risk). Causal
inference, however, is concerned with estimating the causal effect of a change in a covariate.
Regardless, optimal hyperparameters and more data would likely help the SCM work 
better on unseen data. 

Alternatives to the VAE should be further investigated because the causal effect of 
variables on one another is likely unidentifiable in the presence of the latent 
variables $\mathbf{z}$; this means that the functions $f_i$ do not necessarily reflect 
the true causal effect for a change in the parents. 

\Long{
The proposed SCM is a launching point for further refinement. The model is flawed and
almost certainly requires additional variables to properly control for all confounders.
For example, the type of treatment used should be included as variables and would 
make for more interesting counterfactuals (e.g., ``what would this
patient's lesion load be if they had received interferon beta instead of glatiramer acetate?'').
The fact that we don't use this variable in our model can only hamper its performance. 
Better accounting for the relevant variables will improve performance and make more
reliable counterfactual images. 
}

\Short{
The proposed SCM is a launching point but needs further refinement. 
For example, the type of treatment used should be included as 
variables and would make for more interesting counterfactuals (e.g., ``what would this
patient's lesion load be if they had received interferon beta instead of glatiramer acetate?'').
The fact that we do not use this variable in our model can only hamper its performance. 
}

In spite of the limitations, our SCM framework provides a principled way to generate a variety
of images of interest for various applications (e.g., pseudo-healthy image synthesis) and
analysis (e.g., counterfactual analysis of the effect of a medicine on lesion load).
Such a framework can augment normal scientific analysis by providing a principled basis on which
to test the effect of interventions outside of an RCT, and the generated counterfactual images have 
the potential to improve the performance of image processing pipelines and precision medicine in general.

\bibliographystyle{splncs04}
\Long{
\bibliography{biblio}
}
\Short{
\bibliography{biblio_short}
}

\end{document}